\documentclass[letterpaper]{article} % DO NOT CHANGE THIS
\usepackage{aaai24}  % DO NOT CHANGE THIS
\usepackage{times}  % DO NOT CHANGE THIS
\usepackage{helvet}  % DO NOT CHANGE THIS
\usepackage{courier}  % DO NOT CHANGE THIS
\usepackage[hyphens]{url}  % DO NOT CHANGE THIS
\usepackage{graphicx} % DO NOT CHANGE THIS
\usepackage{natbib}  % DO NOT CHANGE THIS AND DO NOT ADD ANY OPTIONS TO IT
\usepackage{caption} % DO NOT CHANGE THIS AND DO NOT ADD ANY OPTIONS TO IT
\usepackage{algorithm}
\usepackage{algorithmic}

\usepackage{newfloat}
\usepackage{listings}
\DeclareCaptionStyle{ruled}{labelfont=normalfont,labelsep=colon,strut=off} % DO NOT CHANGE THIS
\floatstyle{ruled}
\newfloat{listing}{tb}{lst}{}
\floatname{listing}{Listing}
\title{AiGen-FoodReview: A Multimodal Dataset of Machine-Generated Restaurant Reviews and Images on Social Media}
\author{
    Alessandro Gambetti,
    Qiwei Han
}
\affiliations{
    Nova School of Business and Economics, Lisbon, Portugal\\
    Rua da Holanda 1\\
     2775-405 Carcavelos, Lisbon, Portugal\\
    gambetti.alessandro@novasbe.pt, qiwei.han@novasbe.pt
}

\usepackage{bibentry}
\begin{document}

\maketitle

\begin{abstract}
Online reviews in the form of user-generated content (UGC) significantly impact consumer decision-making. However, the pervasive issue of not only human fake content but also machine-generated content challenges UGC's reliability. Recent advances in Large Language Models (LLMs) may pave the way to fabricate indistinguishable fake generated content at a much lower cost.
Leveraging OpenAI's GPT-4-Turbo and DALL-E-2 models, we craft \textbf{AiGen-FoodReview}, a multimodal dataset of 21,143 restaurant review-image pairs divided into authentic and machine-generated. 
We explore unimodal and multimodal detection models, achieving 99.80\% multimodal accuracy with FLAVA. 
We use attributes from readability and photographic theories to score reviews and images, respectively, demonstrating their utility as handcrafted features in scalable and interpretable detection models, with comparable performance.
The paper contributes by open-sourcing the dataset and releasing fake review detectors, recommending its use in unimodal and multimodal fake review detection tasks, and evaluating linguistic and visual features in synthetic versus authentic data.
\end{abstract}

\section{Introduction}

% 1.  what are reviews and images on social media, and their usefulness 

% 2, fake reviews, from human-generated to machine-generated, because of low cost and easy accessibility 

% 3. explain that there are few resources out there now to counteract this phenomenon (this is the research gap). say that this dataset aims to provide a dataset to the open source community to be used for detection and evaluation of fake reviews 

% 4. list tasks of how this dataset can be used: unimodal and multimodal fake review detection (expand). evaluation of deep fake via their attributes (handcrated features) generated. 

% 5. some preliminary findings 

Online reviews on digital platforms are known to significantly influence consumer decisions and trust in products and services. These reviews, integrating real-life experiences, not only offer insights into product quality but also carry economic implications \citep{duan2008online, wu2015economic}. Complementing textual reviews, hybrid reviews that contain both textual and visual content become increasingly prevalent on social media and enhance the richness of user-generated content (UGC), contributing to the overall impact on consumer perception. In particular, there is an emerging research focus on the synergy of texts and images in online reviews. Studies suggest that hybrid content is often perceived as more informative and helpful than textual content \citep{wu2021picture}. For example, research shows that in the context of restaurant reviews, images can have more predictive power for business outcomes than text alone \citep{zhang2023can}.

While online reviews are vital for consumer decision-making, the credibility of UGC is increasingly compromised by the prevalence of fake reviews. These deceptive reviews, whether overly positive or negative, are intended to manipulate consumer perceptions and distort the online marketplace \citep{crawford2015, paul2021}. The restaurant industry, in particular, has seen instances where even marginal increases in online ratings can lead to significant revenue growth, further incentivizing the generation of fake reviews \citep{luca2016reviews}. Platforms such as Amazon, TripAdvisor and Yelp have implemented countermeasures to filter suspicious content. For example, \citet{luca2016fake} found that 16\% of restaurant reviews on Yelp are filtered.

Traditionally, fake reviews have been manually crafted, often by organized fake review farms \citep{he2022market,McCluskey2022}. However, the landscape is evolving with the advancements in Large Language Models (LLMs), which are becoming indistinguishable at producing text from human writing. The accessibility and low cost of these models, such as GPT-based models, pose a new challenge in the proliferation of machine-generated fake reviews \citep{gambetti2023dissecting}. These developments are not limited to text; models like DALL-E are capable of creating realistic images that can accompany fake reviews, adding a new dimension to the challenge of identifying fabricated content \citep{ramesh2021zero}. This situation requires a multifaceted strategy to detect and counteract machine-generated fake content. As visual content manipulation becomes more sophisticated, distinguishing genuine from fabricated imagery is crucial in the broader fight against misinformation. Thus, a comprehensive multimodal approach, encompassing both textual and visual elements, is essential for a robust defense against misinformation in digital spaces.

This paper presents \textbf{AiGen-FoodReview}, a novel multimodal dataset comprising fake restaurant reviews and images, created using OpenAI models GPT-4-Turbo and Dall-E-2. This dataset, with 20,144 review-image pairs, is a pioneering effort in assembling a comprehensive repository of multimodal machine-generated related to customer reviews on social media. We analyze this dataset by examining textual attributes like readability, complexity, and perplexity, as well as visual attributes grounded in photographic theory, such as brightness and compositional elements. Our analysis reveals notable differences between machine-generated and authentic content, with machine-generated reviews displaying higher complexity and the images showing distinct brightness, saturation, and colorfulness.

To address the challenge of separating machine-generated content from authentic material, we developed and optimized several detection models. Our experiments included both unimodal and multimodal machine learning models, focusing on handcrafted features derived from text and images, as well as deep learning models that utilize raw data. In particular, our multimodal FLAVA model, applied to raw data, achieved F1-score of 99.80\%. However, it is worth mentioning that models based on handcrafted features also demonstrated strong performance, offering a viable option when scalability and interpretability are key considerations.

%Here, we showed that machine-generated reviews are more complex in terms of writing, and that machine-generated images tend to be brighter, more saturated, and more colorful. 
%Finally, we optimized several detectors to verify whether machine-generated content is separable from the authentic one. We trained unimodal and multimodal machine learning models on the textual and visual handcrafted features previously computed, and deep learning models on the raw data, showing that multimodal FLAVA on raw data achieves the best performance (99.80\% accuracy). However, machine learning models on handcrafted features achieve comparable performance, factors to be taken into account when scalability and interpretability should be prioritized. 
%This paper has the following contributions. Firstly, we open-source a multimodal dataset of machine-generated fake reviews. Secondly, we 
%This paper has the following contributions. Firstly, we open-source a multimodal dataset of machine-generated fake reviews and images. Secondly, we trained several multimodal fake reviews detectors and we publicly release them to the community. 
This study contributes to the field by introducing a publicly available multimodal dataset of machine-generated fake reviews and images. We also provide a suite of detectors, optimized for identifying such content, and make these tools available to the community. We recommend employing this dataset for the following tasks:

\begin{itemize}
    \item \textbf{Task 1}: Detecting fake restaurant reviews in both unimodal and multimodal contexts.
    \item \textbf{Task 2}: Analyzing and comparing linguistic and visual features of synthetic versus authentic text and image data.
\end{itemize}

%This paper is organized as follows. Firstly, we review the related work for generative AI and the impact of fake reviews in online markets. Next, we present the dataset methodology to construct the final dataset, including a summary statistics.

\section{Related Work}
\subsection{Generative AI and Large Language Models}

Generative AI systems, capable of producing novel content, have evolved significantly in recent years. Early advancements include Generative Adversarial Networks (GANs), which employ a generator
and discriminator in an adversarial training process \citep{goodfellow2014gans}, and Variational Autoencoders (VAEs), focusing on probabilistic mappings \citep{kingma2022autoencoding, mirza2014conditional}. Meanwhile, Pixel Recurrent Neural Networks (PRNN) further extended this concept to sequential image generation \citep{oord2016pixel}. These models essentially laid the groundwork for subsequent developments in content generation. 

The introduction of transformer architecture marked a turning point in generative AI, particularly influencing the development of Large Language Models (LLMs) \citep{vaswani2017attention}. These models have demonstrated remarkable proficiency in producing human-like text and images, reflecting significant progress in the field \citep{chen2021evaluating, kasneci2023chatgpt, koh2023generating}.

LLMs are trained on extensive data to understand and predict patterns, outputting not only indistinguishable human-like text but also generating images via multimodal architectures \citep{chen2021evaluating, kasneci2023chatgpt, koh2023generating}. Unimodal LLMs, exemplified by BERT and RoBERTa, focus primarily on encoding textual information for various predictive tasks. Decoder-focused models like the GPT series excel in generating new text, utilizing a masking strategy to enhance predictive accuracy \citep{devlin2019bert, liu2019roberta, brown2020language, radford2019language}. In contrast, multimodal LLMs integrate different modalities, such as text and images, either through fusion-free or fusion-based architectures. Fusion-free models like CLIP and ALIGN align modality-specific representations via contrastive training, while fusion models like FLAVA and ALBEF use explicit cross-attention mechanisms for intermodal interaction \citep{radford2021learning, jia2021scaling, singh2022flava, li2021albef}.

Moreover, LLMs are increasingly interacted with through prompt engineering, a technique allowing users to guide the generation of responses for specific tasks. This approach has made LLMs like ChatGPT and Bard widely accessible for various applications, including synthetic data generation \citep{white2023prompt, zhou2023large, wu2023brief, veselovsky2023generating, veselovsky2023artificial}. Lastly, LLMs have shown particular effectiveness in synthetic data generation, which includes text production and, more pertinently, the generation of fake reviews. The democratization of these technologies raises concerns about their potential misuse in creating deceptive online content \citep{gambetti2023dissecting}.

%% say something on syntehtic data generation, bridge to fake reviews on social media 

%\subsection{Fake Review Datasets}

\subsection{Fake Reviews on Social Media and Online Markets}

The proliferation of fake reviews on social media and online markets has become a significant issue. Motivated by monetary gains, retailers and online platforms manipulate reviews to influence consumer behavior, affecting market dynamics and trust \citep{crawford2015, gossling2018manager, lee2018sentiment, paul2021, he2022market}.
Fake reviews, by distorting the informativeness of content, have profound implications on consumer decision-making. They foster uncertainty and distrust among consumers, negatively affecting their purchasing intentions \citep{agnihotri2016online, zhao2013modeling, zhang2017welfare, deandrea2018people, filieri2015travelers, munzel2016assisting, xu2020effects, zhuang2018manufactured}.

An emerging concern is the use of LLMs for generating fake reviews. The ability of these models to produce authentic-seeming content at scale poses new challenges for online platforms in identifying and mitigating such deceptive practices. This development calls for advanced detection methods and a reevaluation of current strategies to preserve the integrity of online review ecosystems \citep{gambetti2023dissecting}. As machine-generated content becomes more sophisticated, the detection of fake reviews requires not only traditional text analysis but also an understanding of the nuances introduced by AI-generated content. This underscores the need for continuous research and development in AI detection methodologies, emphasizing
the importance of both unimodal and multimodal approaches to effectively identify synthetic content.

%say something on fake reviews generated by machines 
% say something on the applications of large language models 

\section{Dataset Methodology}
\subsection{Raw Data Collection}
We leveraged the New York City SafeGraph restaurant mobility data from 2019 to June 2022 as a basis for selecting restaurants to collect reviews (\url{https://www.safegraph.com/}). SafeGraph is a company that tracks customer traffic of point-of-interest (POIs) such as restaurants via smartphone signals (Wi-Fi, GPS, and Bluetooth). We selected New York City as a geographical area as it offers a diverse set of international cuisines in a global setting, ensuring cultural heterogeneity for our research. 

\begin{figure}[t]
\centering
\includegraphics[width=0.9\columnwidth]{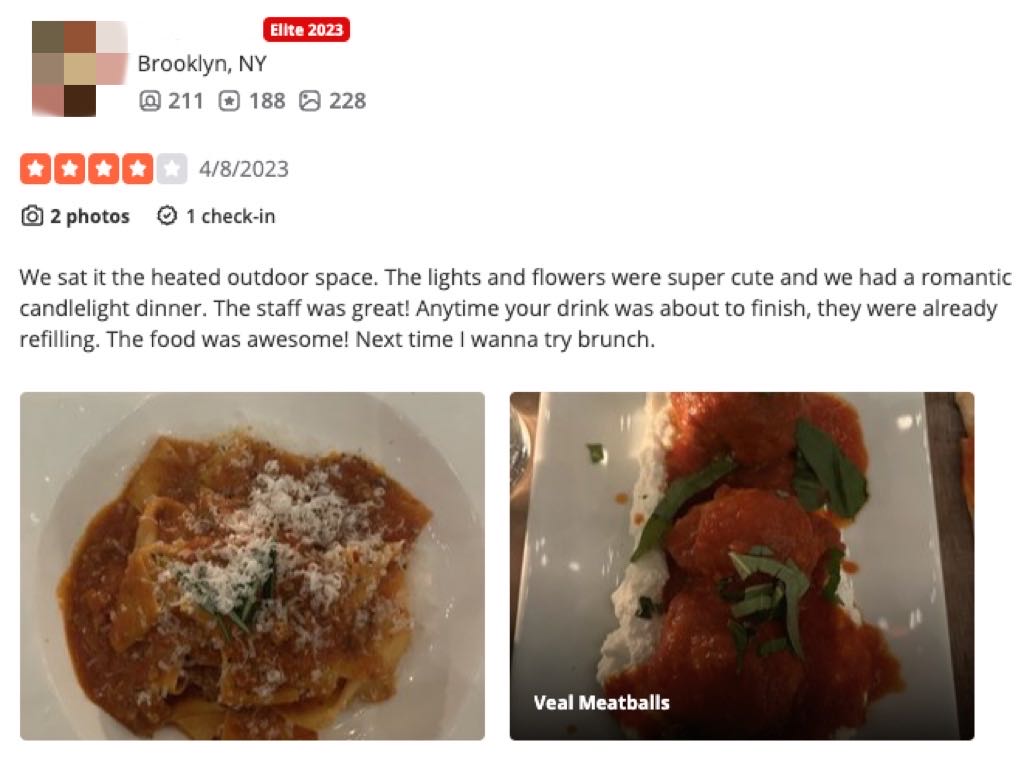} % Reduce the figure size so that it is slightly narrower than the column. Don't use precise values for figure width.This setup will avoid overfull boxes.
\caption{Example of how a scraped review (with images attached by the same user) is displayed on Yelp. The red label indicates the user's elite status. Other variables include user location, user number of friends, number of previous reviews posted, and number of images posted. Name and image were anonymized and blurred.}
\label{fig:example_review}
\end{figure}

From the initial set of 9,200 restaurants, we scraped all English-written reviews from Yelp, a prominent platform for discovering and reviewing local businesses. This effort yielded 447,377 textual reviews from the same period. Alongside each review, we collected associated ratings, user elite status, and downloaded accompanying images, if any. The relationship between reviews and images was carefully mapped, with each review potentially linked to multiple images. Figure \ref{fig:example_review} shows an example of a scraped review displayed on Yelp. 

%\vspace{-0.35cm}
\subsection{Data Processing}
Our processing involved two key steps. Initially, we focused on reviews with at least one attached image, ensuring a multimodal dataset. For reviews linked to multiple images, we randomly sampled a single image to pair with the text. This random sampling was crucial to maintain diversity and avoid bias in image representation. We then narrowed down to reviews posted by Yelp's elite users, considering these as more reliable sources. This choice is made with two considerations: firstly, elite status on Yelp is a mark of verified and consistent contribution, indicative of trustworthy reviews \citep{zhang2020matter, wang2021examining}; secondly, Yelp's filtering system, while effective, may not fully capture all fake reviews \citep{mukherjee2013yelp}, and selecting only elite reviews adds an additional layer of authenticity. Consequently, we propose that elite reviews offer the closest proxy to genuine customer feedback. Finally, the textual data was cleaned to remove HTML tags and non-ASCII characters, resulting in a dataset of 21,143 elite reviews, each paired with at least one image.

\subsection{Multimodal Fake Reviews and Images Generation}
\begin{figure}[t]
\centering
\includegraphics[width=0.9\columnwidth]{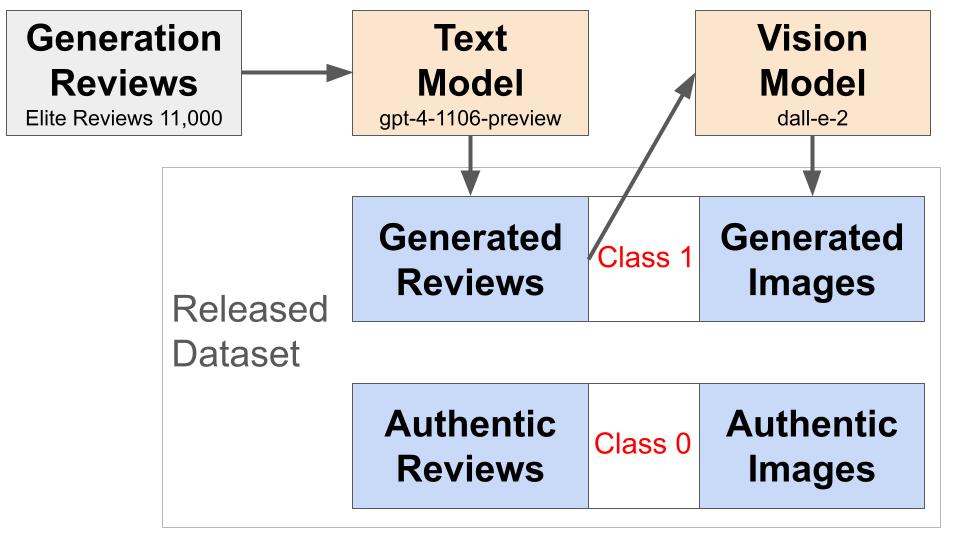} % Reduce the figure size so that it is slightly narrower than the column. Don't use precise values for figure width.This setup will avoid overfull boxes.
\caption{Diagram of the data generation methodology. Elite reviews are used as a source of information in the prompt to query the GPT-4-Turbo model to generate a fake review. Next, the \textit{generated} fake review is used as information to produce a related synthetic image. Finally, review-image pairs, both \textit{authentic} and \textit{generated}, were aggregated into the final dataset to form the negative class (\textit{Class 0}) and positive class (\textit{Class 1}), respectively.}
\label{fig:methodology}
\end{figure}

To generate the multimodal dataset, we employed GPT-4-Turbo for text generation \citep{openai2023} and Dall-E-2 for image creation \citep{ramesh2022hierarchical}. These state-of-the-art models
were selected for their advanced capabilities in generating realistic and contextually relevant content. The dataset generation process involved several carefully designed steps, as illustrated in Figure \ref{fig:methodology}. Importantly, we define the following terminology: (1) \textit{generation} are the reviews used as input as the source of information to give contextual knowledge to the language model to generate synthetic text, (2) \textit{generated} refers to the generated output from the language and vision models (both reviews and images), and (3) \textit{authentic} refers to the reviews and images aggregated as ground truth later. Figure \ref{fig:methodology} shows a visual summary diagram of the data generation methodology. The steps are explained as follows.

Firstly, we randomly divided the 21,143 reviews between 11,000 examples to be used for \textit{generation} and 10,143 as \textit{authentic} to create a binary dataset between the two classes. 
It is reasonable to separate \textit{generation} data and data to be aggregated later as \textit{authentic} because of self-containment. For example, assuming separation had not been performed, there would be a probability that \textit{generated} reviews would be similar to \textit{generation} reviews. Therefore, in a train-test split scenario, it could be likely that a detector would learn wrong representations.
%In other words, \textit{generated} and \textit{generation} reviews could occur in the train and test split separately, rendering the learning process difficult to accomplish.
%The rationale behind separating data is to heterogeneously represent the likelihood of coming across a machine-generated revi
Secondly, for each review in \textit{generation}, we queried the \textit{gpt-4-1106-preview} model with the following prompt, which corresponds to the default prompt embedded into the fake review generator tool that OpenAI provided in its playground as of early 2023 \citep{gambetti2023dissecting}%(https://platform.openai.com/examples/default-restaurant-review):
\footnote{\url{https://platform.openai.com/examples/default-restaurant-review}}:
\newpage

{\centering
\textit{``Write a restaurant review based on these notes:}\\
\textit{Name: \textless EXAMPLE RESTAURANT NAME\textgreater}\\
\textit{\textless EXAMPLE ELITE REVIEW TEXT\textgreater"} 
\par}
Other hyperparameters such as the \textit{temperature} and the \textit{top\_p} were kept as default, except for the \textit{max\_length} of the generated output set to 512 tokens to avoid output truncation. 

\begin{figure}[t]
\centering
\includegraphics[width=0.9\columnwidth]{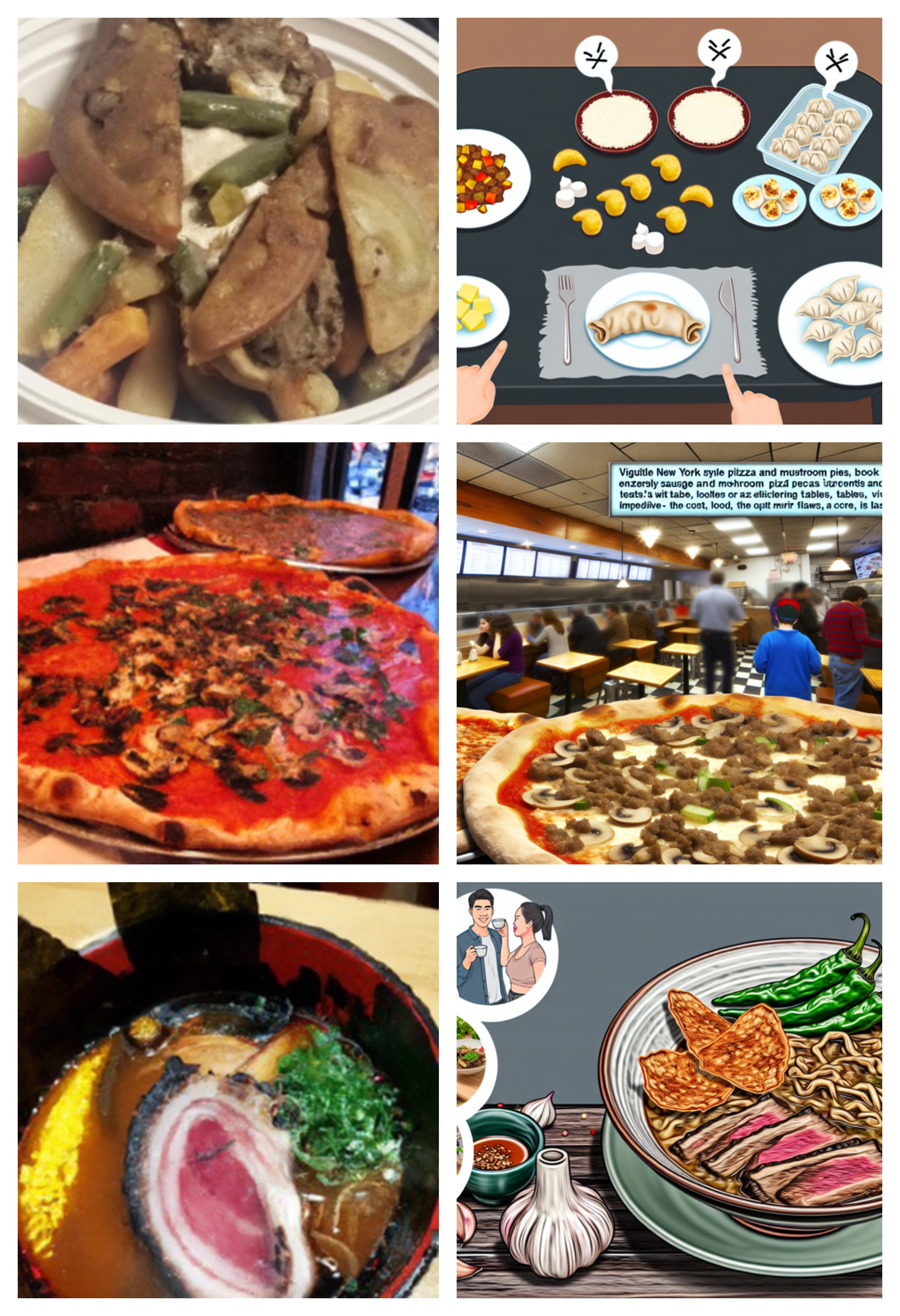} % Reduce the figure size so that it is slightly narrower than the column. Don't use precise values for figure width.This setup will avoid overfull boxes.
\caption{Generated Images from Dall-E-2 (left column) and Dall-E-3 (right column) for the same prompt.}
\label{fig:sample_generated_images}
\end{figure}

Thirdly, we prompted the \textit{dall-e-2} standard model with the \textit{generated} reviews to generate synthetic images of size 256x256 based on synthetic text. All the hyperparameters were kept as default. 
Since \textit{dall-e-2} accepts a maximum of 1,000 characters, prompt truncation was performed at that cutoff value. The reason we used \textit{dall-e-2} instead of the most recent \textit{dall-e-3} is two-fold: (1) \textit{dall-e-3} is 2.5 times more costly than \textit{dall-e-2} with prices of \$0.040/image (standard model, size 1024×1024) and \$0.016/image (standard model, size 256x256), respectively, and (2) we noticed from our experiments that \textit{dall-e-3} outputs less realistic and more cartoonish images than \textit{dall-e-2}. We validated that by prompting \textit{dall-e-3} with the same hyperparameters as for \textit{dall-e-2} but adding \textit{style} equal to \textit{natural} to avoid generating hyper-real images. In Figure \ref{fig:sample_generated_images} we show 3 samples of generated images from the two models using the same prompt out of a total of 30 pairs generated. Validation was done manually by the authors. 

% talk about the errors with code 
Finally, because of content moderation imposed by OpenAI, we recorded 999 errors in generating images, meaning that 999 \textit{generated} reviews could not be ``translated" into images (Error code: 400, ``content\_policy\_violation"). 
Thus, we aggregated to the 10,001 \textit{generated} review-image pairs (\textit{Class 1}) the 10,143 \textit{authentic} reviews and related images (\textit{Class 0}). In total, the final multimodal review dataset counts 20,144 review-image pairs.

\subsubsection{Generation Cost Breakdown.}
We here report the cost incurred for the steps illustrated above. As of January 2024, \textit{gpt-4-1106-preview} prices input text at \$0.01/1K tokens and output text at \$0.03/1K tokens. Here, we counted the average number of input and output tokens using the popular Python package ``tiktoken". 
Whereas, \textit{dall-e-2} (standard 255x256) prices a single image generation at \$0.016. 
Arithmetically, we can break the \textit{TOTAL} cost as:

\begin{equation}
  \begin{array}{ll}
    TC = (0.01 * IC + 0.03 * OC) * M \\ 
    VC = 0.16 * M \\
    TOTAL = TC + VC
  \end{array}
\end{equation}

\textit{TC} stands for the text cost using a textual model, \textit{VC} stands for the image cost, $IC=188$ stands for the average number of input tokens (average length in \textit{generation}), and $OC=157$ stands for the average number of output tokens (average length in \textit{generated}). Finally, $M=11,000$ is the number of \textit{generation} reviews. In total, $TC$ amounts to about \$72, while $VC$ amounts to about \$176, for a \textit{TOTAL} cost of about \$248. 

It is important to highlight that machine-generated reviews are relatively cheap to generate, as one machine-generated review costs, on average, about \$0.00654 ($TC / M$). For the sake of comparison, one investigation from Which, a UK consumer-right group,  reported the cost of buying human-generated fake reviews at \$18 per review \citep{dean2021reviews}. 
%For this reason, we conjecture that machine-generated fake reviews may proliferate soon because of their low cost. 

\subsection{Dataset Summary Statistics} \label{sec:eda}
We provide a concise dataset summary statistics at review, image, and restaurant levels for the restaurants represented in \textit{authentic} and in \textit{generated}. As for terminology clarification, we interchangeably refer to the \textit{authentic} reviews as \textit{Class 0}, and to the \textit{generated} reviews as \textit{Class 1}, respectively. 

\subsubsection{Review Level.}
The average rating (stars, 1:5), represented both in \textit{generated} and \textit{authentic}, amounts to 4.07 (std 0.90), meaning that most of the reviews represented were positive overall. Also, no rating differences across \textit{Class 0} and \textit{Class 1} were recorded, with average ratings of 4.074 (std 0.91) and 4.073 (std 0.90), respectively. 

% decide whether include this or not 
%Then, \textit{authentic} reviews are (on average) as long as \textit{generated} reviews, with averages of 151 (std 102) and 157 (std 57) words per review, respectively. 

Next, we calculated review text statistics by mining readability, complexity, and perplexity handcrafted attributes. Here, we scored each review with the (1) Automated Readability Index (ARI), which measures an approximate representation of the US grade level needed to comprehend the text \citep{senter1967automated}; (2) the Fleisch Readability index (FR), in which higher scores indicate text that is easier to read \citep{flesch1948new}; (3) the number of difficult words (DW) present in the Dale-Chall word list, which approximately contains 3,000 words of difficult understanding \citep{dale1948formula}; (4) the Gunning Fog Index (GFI), which has a similar meaning as for ARI; (5) the reading time (RT), which calculates the reading time assuming 14.69ms per character; (6) the average words per sentence (WPS); 
and (7) with perplexity (PPL). Here, for each review, we calculated PPL as the exponential weighted average of the negative log-likelihoods of word sequences. To do so, we implemented a zero-shot 125M parameters GPTNeo model, which is an open-source replication of the GPT-3 model \citep{black2022gptneox20b}. We utilized ANOVA to assess the statistical differences among variables.
In Table \ref{tab:variable_description}, we describe the variables, and in Table \ref{tab:text_summary_stats} we report the text summary statistics.

\begin{table*}[t]
\centering
%\resizebox{.95\columnwidth}{!}{
\begin{tabular}{l|l}
\hline
 \textbf{Metric} & \textbf{Description} \\
    \hline
    ARI  & Higher ARI, more difficult text: output is the US grade school level required for comprehension. \\
    FR & Score on a scale from 0 to 100, with higher scores indicating easier readability.  \\
    DW & List of 3,000 common words: if a word is not on the list, it is considered difficult. \\
    PPL & Lower perplexity suggests lower text entropy and better predictability: tokens are predictable. \\
    GFI & Years of formal education a person needs to understand a text easily. \\
    RT & Reading times. \\
    WPS & Average number of words per sentence. \\
    \hline
    FA & Food aesthetics score from a food aesthetic model trained by \citet{gambetti2022camera}. \\
    BRI & Brightness. Average of V of the HSV image representation. \\
    SAT &  Saturation. Color intensity and
    purity of an image. Average of S of the HSV image representation.\\ 
    CON & Contrast. Spread of illumination. Standard deviation of V of the HSV image representation. \\
    CLA & Clarity. Well-defined objects in space. \% of normalized V pixels that exceed 0.7 of HSV. \\
    WAR & Warmth. Warm colors: from red to yellow. \% of H $<$
    60 or $>$ than 220 of HSV. \\
    COL & Colorfulness. Departure from a grey-scale image. \\
    SD & Size difference. Difference in
    the number of pixels between the figure and the ground. \\
    CD & Color difference. Difference
    of Euclidian distance between the figure and ground (RGB vectors). \\
    TD & Texture difference. Absolute difference between the foreground and background edge density. \\
    DD & Diagonal dominance. Manhattan distance between salient region and each diagonal. \\
    ROT & Rule of thirds. Minimum distance between center of salient region and each of the four intersection points. \\
    HPVB & Horizontal physical visual balance. Split image horizontally. Horizontal physical symmetry (mirroring). \\
    VPVB & Vertical physical visual balance. Split image vertically. Vertical physical symmetry (mirroring). \\
    HCVC & Horizontal color visual balance. Split image horizontally. Horizontal mirrored Euclidean cross-pixels distance. \\
    VCVC & Vertical color visual balance. Split image vertically. Vertical mirrored Euclidean cross-pixels distance. \\
    \hline
    
\end{tabular}
\caption{Description of the textual and visual handcrafted variables mined in the paper.}
\label{tab:variable_description}
\end{table*}

\begin{table}[t]
\centering
%\resizebox{.95\columnwidth}{!}{
\begin{tabular}{l|l|l|l}
\hline
 \textbf{Metric} & \textbf{Authentic} & \textbf{Generated} & \textbf{F-statistic} \\
    \hline

    ARI  & 6.84 (3.00) &  12.20 (1.90) & 22,883*** \\
    FR  &  79.93 (9.26)  & 57.78 (8.15) & 32,424*** \\
    DW  & 19.77 (13.97) & 39.06 (15.23) &  8,783*** \\
    PPL  &  55.56 (41.43)  & 38.19 (11.31) & 1,638*** \\
    GFI & 7.71 (2.37) & 12.29 (1.66) & 25,309*** \\
    RT & 9.93 (6.82) & 11.96 (4.44) & 626*** \\
    WPS & 14.10 (5.39) & 19.05 (3.15) & 6,303*** \\
    \hline

\end{tabular}
\caption{Summary of text statistics. Average values are reported with standard deviation in brackets. *\textit{p}$<$.05, **\textit{p}$<$.01, ***\textit{p}$<$.001.}
\label{tab:text_summary_stats}
\end{table}

Results indicate that: \textit{generated} reviews are more complex to read as compared to \textit{authentic} ones, as measured by higher ARI and GFI, lower FR, and a larger number of DW. Also, \textit{generated} reviews score an average lower perplexity, which is coherent with LLMs optimization logic, as such models are generally optimized by minimizing perplexity in the self-supervised learning paradigm. Finally, \textit{authentic} reviews have longer sequences (higher WPS), which translates into longer reading time (RT).

\subsubsection{Image Level.} 
We scored each image with the attributes illustrated by \citet{gambetti2022camera} to calculate a statistics of photographic attributes per class, including their developed food aesthetic score for food images only (FA).
To do so, we sampled 20,000 labeled images from the Yelp official dataset\footnote{https://www.yelp.com/dataset}, split the data into 80\% train, and 20\% test, and fine-tuned a ViT-B/16 model to classify food versus non-food images, achieving 97.63\% accuracy on the test set. Finally, we applied FA to food images. FA ranges from 0 (low aesthetics) to 1 (high aesthetics).

Next, we considered the following photographic attributes: (1) color attributes such as brightness (BRI), saturation (SAT), contrast (CON), clarity (CLA), warmth (WAR), and colorfulness (COL); (2) figure-ground relationship attributes such as size difference (SD), color difference (CD), and texture difference (TD); and (3) image composition attributes such as diagonal dominance (DD), rule of thirds (ROT), horizontal/vertical physical visual balance (HPVB, VPVB), and horizontal/vertical color visual balance (HCVB, VCVB). We utilized ANOVA to assess the statistical differences among variables.
In Table \ref{tab:variable_description} we describe the variables, and in Table \ref{tab:image_summary_stats} we show the summary statistics.

\begin{table}[t]
\centering
%\resizebox{.95\columnwidth}{!}{
\begin{tabular}{l|l|l|l}
\hline
 \textbf{Metric} & \textbf{Authentic} & \textbf{Generated} & \textbf{F-statistic} \\
    \hline
    FA & 0.07 (0.15) & 0.35 (0.34) & 4,604***  \\
    \hline
    BRI  & 132.79 (30.41) &  144.51 (33.73) & 671*** \\
    SAT  &  112.65 (35.79)  & 117.35 (42.45) & 1,822***  \\
    CON  &  59.16 (11.66) &  69.81 (12.48) & 3,919***  \\
    CLA  & 0.30 (0.18)  & 0.41 (0.18) & 1,822*** \\
    WAR  & 0.28 (0.21) & 0.45 (0.21) & 3,183***  \\
    COL  & 150.11 (18.67) & 160.88 (14.49) & 2,089*** \\
    \hline
    SD & 0.31 (0.29) &  0.29 (0.32) & 18.74***\\
    CD & 79.78 (52.21) & 93.94 (57.78) & 332***\\ 
    TD & 4.06 (11.11) & 3.87 (4.64) & 2.69  \\
    \hline
    DD & -32.76 (33.77) & -36.92 (33.52) & 76.78*** \\
    ROT &  -75.65 (23.29) & -79.67 (15.16) & 209.67***\\ 
    HPVB & -9.09 (8.11) & -9.00 (7.46) & 0.78\\ 
    VPVB & -6.41 (5.77) & -6.36 (5.34) & 0.40\\ 
    HCVB & -0.54 (0.14) & -0.48 (0.02) & 1,977*** \\ 
    VCVB & -0.54 (0.15) & -0.48 (0.03) & 1,598*** \\ 
    \hline

\end{tabular}
\caption{Summary statistic of image attributes. Average values are reported with standard deviation in brackets. *\textit{p} $<$.05, **\textit{p}$<$.01, ***\textit{p}$<$.001.}
\label{tab:image_summary_stats}
\end{table}

Results indicate that: \textit{generated} images tend to score a higher aesthetics score (FA). As for color attributes, \textit{generated} images are brighter (higher BRI), more saturated (higher SAT), clearer (higher CLA), warmer (higher WAR), and more colorful (higher COL). Next, as for figure-ground relationship, both \textit{authentic} and \textit{generated} images show comparable size differences (SD) and texture differences (TD) between the figure and the ground. However, \textit{generated} images tend to be more separated in terms of color difference between foreground and background (CD). Finally, although statistically significant, as for image composition, no clear patterns could be evinced across the two classes, showing comparable diagonal dominance (DD), rule of thirds (ROT), physical visual balance (HPVB, VPVB), and color visual balance (HCVB, VCVB).

\subsubsection{Restaurant Level.}
There are 3,238 restaurants represented in the dataset, with an average of about 6 reviews per restaurant. We then queried the official Yelp Fusion API to gather information about the regional cuisines represented in the sample (\url{https://fusion.yelp.com/}). In Table \ref{tab:regional_cuisines}, we show the top 10 regional cuisines.
\begin{table}[t]
\centering
%\resizebox{.95\columnwidth}{!}{
\begin{tabular}{l|l|l}
\hline
    \textbf{Cuisines} & \textbf{\#Restaurants} & \textbf{\%Restaurants} \\
    \hline
    American & 610 & 18.84 \\
    Italian & 428 & 13.22 \\
    Japanese & 187 & 5.78\\ 
    Chinese & 186 & 5.74\\
    Mexican & 165 & 5.10 \\ 
    French & 141 &  4.35 \\ 
    Thai & 128 & 3.95 \\
    Mediterranean & 127 & 3.92 \\
    Korean & 96 & 2.96 \\
    Indian & 88 & 2.72 \\
    \hline
\end{tabular}
\caption{List of top 10 regional cuisines represented in the dataset. The number of restaurants is reported both in absolute and relative number of occurrences.}
\label{tab:regional_cuisines}
\end{table}
American, Italian, and Japanese are the most popular cuisines with market shares of 18.84\%, 13.22\%, and 5.78\%, respectively. Next, in terms of price levels, which indicate the average meal price per person, 1,941 restaurants (59.94\%) are priced in the range \$11–\$30, as denoted by \$\$. In descending order, 452 restaurants are priced between \$31–\$60 (\$\$\$), 372 restaurants are priced below \$10 (\$), and 119 over \$61 (\$\$\$\$).
Finally, the average of the restaurants' average rating amounts to 3.80 (std 0.51), which is minimally lower than the average rating of \textit{Class 0} and \textit{Class 1} reviews.

\section{Experiments}
\begin{figure}[t]
\centering
\includegraphics[width=0.9\columnwidth]{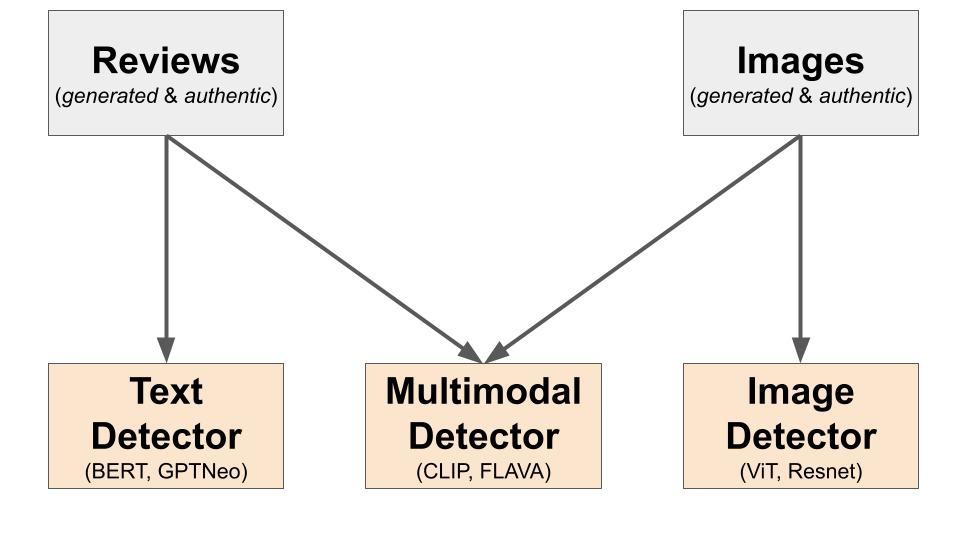} % Reduce the figure size so that it is slightly narrower than the column. Don't use precise values for figure width.This setup will avoid overfull boxes.
\caption{Representation of unimodal and multimodal feature combinations for detection models.}
\label{fig:detection}
\end{figure}

We trained unimodal and multimodal binary fake review detectors to test how they perform on our crafted dataset. Firstly, we randomly split the generated dataset into 60\% train, 20\% validation, and 20\% test. Secondly, we trained unimodal models on text and image data separately, and finally trained multimodal models on text and image data together (see Figure \ref{fig:detection}).
We also report the performance of open-source models as benchmarks. Evaluation metrics for comparison include the accuracy score, precision score, recall score, and F1-score. 

\subsection{Text Detectors}
We benchmarked with the official OpenAI RoBERTa model for fake text detection \citep{solaiman2019release}. We applied it directly to the test set. 
Next, we fine-tuned a 110M parameters BERT and a 125M parameters GPTNeo \citep{black2022gptneox20b} for binary classification. 
We trained with AdamW with a learning rate of 1e-4, a batch size of 16, truncation and padding at 512 tokens, and early stopping at 5 epochs. We trained on the training set and evaluated accuracy after each epoch on the validation set. Finally, we reported the results of the test set. 

\subsection{Image Detectors}
By directly applying it to the test set, we benchmarked with an open-source CvT-13 model trained to detect AI-generated images by \citet{horbatko2023}, who optimized on ArtiFact \citep{rahman2023artifact}, a generalist dataset for synthetic image detection. Then, we fine-tuned a ViT-B/16 and a ResNet-50. We trained with equal methodology to text models: AdamW with a learning rate of 1e-4, batch size of 16, and early stopping at 5 epochs. We followed the default image-processing steps for both models in their original papers.

\subsection{Multimodal Detectors}
To the best of our knowledge, no robust benchmarks exist for text-image multimodal deepfake detection. Thus, we could not benchmark prior models with our dataset. Hence, we fine-tuned pre-trained CLIP (based on a ViT-B/16), a fusion-free multimodal model \citep{radford2021learning}; and FLAVA, a late-fusion multimodal model \citep{singh2022flava}. 
Both were pre-trained by learning representations from image and text pairs through contrastive training.
As for CLIP, we extracted the separate image and text-learned representations and concatenated them. On top, a linear MLP was added as a classification head. As for FLAVA, we added a linear MLP on the jointly learned representation at the last layer as a classification head. Without freezing the backbone models, we trained with the same hyperparameters and methodology as for unimodal models: AdamW with a learning rate 1e-4, a batch size of 16, truncation at 512 tokens (except for CLIP, which truncates inputs at 77 tokens \citep{urbanek2023picture}), and early stopping at 5 epochs.  

\subsection{Handcrafted Features Detectors}
We evaluated whether handcrafted textual and image features contribute to detecting machine-generated reviews. Using the features mined in the \textit{Dataset Summary Statistics} section, we optimized Logistic Regression models (LR) and Random Forest (RF) on: (1) text features only, \textit{i.e.}, readability, complexity, and perplexity features, (2) image features only, \textit{i.e.}, photographic attributes, and (3) a multimodal aggregation including both sets of features. We trained on the training set, and reported results on the test set. Hyperparameters were kept as default.
Results are reported in Table \ref{tab:experiment_results}.
Finally, given the explainability nature of RF and the handcrafted features, we applied SHAP on the test set to check the feature importance of the most impactful features in making predictions. 
For general reference, SHAP (SHapley Additive exPlanations) is a widely used interpretability framework in machine learning, offering a unified and theoretically grounded approach to quantify the contribution of each feature in a model's output \citep{lundberg2017shap}.

\subsection{Results \& Discussion}
\begin{figure}[t]
\centering
\includegraphics[width=0.9\columnwidth]{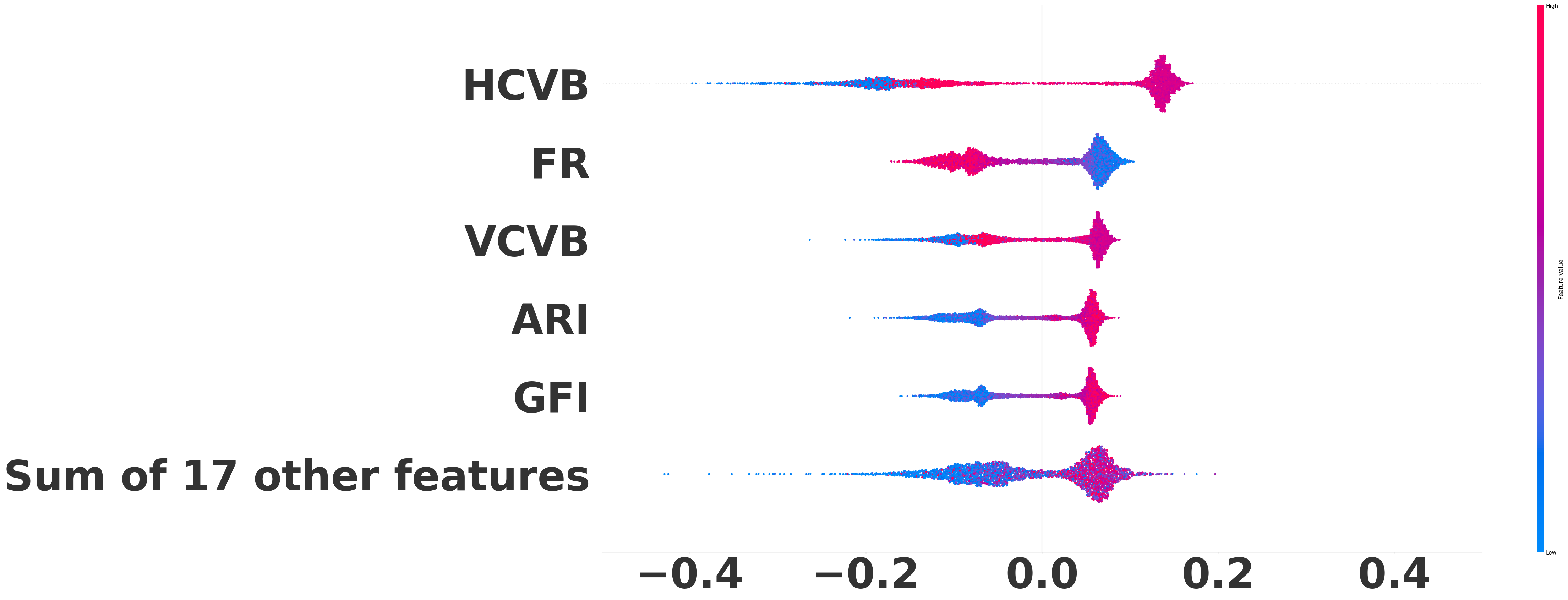} % Reduce the figure size so that it is slightly narrower than the column. Don't use precise values for figure width.This setup will avoid overfull boxes.
\caption{SHAP evaluation: top 5 influential features to predict \textit{generated} reviews in descending order. The concentration of red dots in the $x>0$ quadrant, with the blue ones on the $x<0$ quadrant, implies a positive correlation with the target variable, and vice versa. For example, ARI is positively correlated, while FR is negatively correlated.}
\label{fig:shap}
\end{figure}

We summarize the binary classification report on the test set in Table \ref{tab:experiment_results}. 
\begin{table*}[t]
\centering
%\resizebox{.95\columnwidth}{!}{
\begin{tabular}{l|l|l|l|l|l|l}
\hline
    \textbf{Model} & \textbf{Type} & \textbf{Conv@Epoch}& \textbf{Accuracy \%} &    \textbf{Precision \%} & \textbf{Recall \%} & \textbf{F1-score \%}  \\
    \hline
    LR (ours) & Handcrafted/Text & - & 94.99 & 95.14 & 94.96 & 95.05 \\
    RF (ours) & Handcrafted/Text & - & 95.46 & 95.81 & 95.21 & 95.51 \\
    LR (ours) & Handcrafted/Image & - & 78.24 & 78.62 & 78.42 & 78.52 \\
    RF (ours) & Handcrafted/Image & - & 98.19 & 98.24 & 98.19 & 98.21 \\
    LR (ours) & Handcrafted/Multi & - & 96.30 & 96.65 & 96.03 & 96.34 \\
    RF (ours) & Handcrafted/Multi & - & 98.96 & 99.31 & 98.63 & 98.97 \\

    \hline
    \citet{solaiman2019release} & Text & - & 44.94 & 22.57 & 03.52 & 06.09 \\
    BERT (ours) & Text & 1 & 99.38 & 99.51 &  99.27 & 99.39 \\
    GPTNeo (ours) & Text & 6 & 99.68 & 99.51  &  \textbf{99.85} & 99.68 \\
    %\hline
    \citet{horbatko2023} & Image & - & 73.67 & 88.01 & 68.80 & 77.23 \\
    ViT-B/16 (ours) & Image & 3 & 99.16 & 99.46 & 98.88 & 99.17 \\
    ResNet-50 (ours) & Image & 8 & 99.21 & \textbf{99.90} & 98.55 & 99.22 \\
    %\hline
    CLIP (ours) & Multimodal & 20  & 98.46 & 98.34 & 98.63 & 98.48 \\
    FLAVA (ours) & Multimodal & 1  & \textbf{99.80} & \textbf{99.90} & 99.71 & \textbf{99.80} \\
    \hline

\end{tabular}
\caption{Classification report on the test set. Numbers in bold indicate best performance. Conv@Epoch indicates the epoch in which weights were saved after early stopping.}
\label{tab:experiment_results}
\end{table*}
Results indicate that restaurant \textit{generated} content is separable from \textit{authentic} content both in unimodal and multimodal experiments. 
Benchmarks' performance is not sufficient to achieve satisfactory accuracy. Here, the official OpenAI RoBERTa detector performed worse than random guessing (\textless50\%) in the unimodal text setting, and \citet{horbatko2023} vision model achieves only 73.67\% accuracy in the unimodal image setting. In contrast, all the fine-tuned models achieved about 99\% performance in all the metrics evaluated.  
With an accuracy comparable to BERT (99.38\%), GPTNeo achieves the best accuracy and F1-score for text models with values of 99.68\% and 99.68\%, respectively. As for unimodal image models, ResNet-50 has comparable accuracy to ViT-B/16 (99.21\% and 99.16\%, respectively). 
As for multimodal models, FLAVA outperformed CLIP by +1.34\% with an accuracy of 99.80\%, and all the other unimodal models. 
Finally, as for models trained on handcrafted features, RF trained on multimodal features achieves the best performance (98.96\% accuracy). This is followed by the RF on image features (98.19\% accuracy) and LR on multimodal features (96.30\% accuracy). We then applied SHAP explainability to the multimodal RF model. Figure \ref{fig:shap} shows the top 5 influential features in predicting a \textit{generated} review-image pair. From these results, both horizontal and vertical color visual balance (HCVB, VCVB) have a higher weight, as well as Flesch Reading index (FR), Automated Readability Index (ARI), and Gunning Fog Index (GFI).

Overall, all the fine-tuned models, except for LR trained on handcrafted image features, achieved robust performance in the hold-out dataset. However, multimodal FLAVA achieves the overall best performance both in accuracy and F1-score, suggesting that detection from multimodal input can be an effective alternative when possible. However, CLIP performed worse, conjecturing that CLIP's forced truncation at 77 textual input tokens can be a limiting factor during optimization \citep{urbanek2023picture}. For reference, input for the other text models was truncated at 512 tokens. Inherently, CLIP's convergence after 20 epochs might be a consequence stemming from this limitation. 
A valid alternative to deep learning models was the adoption of standard machine learning models trained on handcrafted features. The latter achieved comparable performance, but with faster training and inference time, enabling cost savings and quicker experimentation.
These factors should be taken into account in real-world scenarios in which scalability plays a central role in project development. Also, training machine learning models on handcrafted features enhances their interpretability and explainability, facilitating a clearer understanding of the factors influencing their predictions. In this regard, we showed that SHAP is a valuable framework for providing explainability to a machine-generated fake reviews detector based on handcrafted features. 
%Lastly, we open-source all the optimized weights and provide examples of how the model could be used at \url{INSERT URL OF GITHUB REPO}.

\subsection{Limitations}
This work is not without limitations. Firstly, we relied on New York City to collect data to avoid sampling bias. Thus, NY is the only geographical area represented in the dataset. Secondly, we only adopted one generative model for each task: GPT-4-Turbo and Dall-E-2 for text and vision, respectively. Such models are the most popular ones in terms of user adoption. However, alternative LLMs were not used, making the dataset suitable for mostly detecting OpenAI-generated content. Thirdly, we only adopted one prompt template for both text and image generation. Here, spammers may use a variety of prompts, resulting in varied complexity of the generated content.

\section{Conclusion}

In this paper, we have presented \textbf{AiGen-FoodReview}, a multimodal dataset of machine-generated reviews and images. The dataset comprises 20,144 review-image pairs, divided into 10,143 \textit{authentic} and 10,001 \textit{generated} pairs. \textit{Authentic} reviews were scraped from Yelp, an online platform where users can find and share reviews about local businesses, while \textit{generated} reviews and images were crafted by prompting GPT-4-Turbo and Dalle-E-2, respectively, for a total cost of about \$248, which is significantly lower than those incurred by hiring human workers to do the task (see \citet{dean2021reviews}). This finding underscores the need for further investigation in future research on whether machine-generated content may become more prevalent with respect to human-generated content.
We then performed an exploratory data analysis by comparing \textit{authentic} and \textit{generated} data in terms of handcrafted review level, image level, and restaurant level variables, primarily showing that GPT-4-Turbo \textit{generated} reviews are more linguistically complex, and that Dall-E-2 \textit{generated} images are brighter, more saturated, and clearer as compared to the authentic ones. 
Next, we optimized several unimodal and multimodal fake review-image detectors with handcrafted features and with raw unstructured data as input, showing that all achieved reasonable performance on the test set. 
With the mass adoption of LLMs and their easy accessibility, we conjecture that machine-generated reviews could proliferate in the near future, distorting user experiences, and undermining trust in social media platforms. 
In conclusion, this dataset is a tentative to provide the open-source community with a research and educational benchmark to develop knowledge on multimodal machine-generated reviews and images. 

% 1. short review of what has been done
% 2. list of the tasks where the dataset can be used

\section{Ethical Impact and FAIR}
% 1. upload the dataset somewhere wwere they recommend and state where you can find it 

This dataset release aligns with ethical standards to ensure integrity in research. Privacy and confidentiality have been rigorously maintained through anonymization and securing of informed consent. Therefore, no private information about Yelp users has been disclosed. We release the dataset with an MIT license. Researchers are urged to engage in the responsible utilization of this dataset, as well as not using the content of this dataset to spread misinformation on the web. 
We strongly encourage the academic community (and non) to leverage this resource conscientiously.

We adhere to the FAIR principles. The dataset is \textbf{findable} by its hosting on Zenodo, accompanied by a unique DOI identifier: 10.5281/zenodo.10511456. This ensures that the dataset is easily discoverable and identifiable. The dataset files are \textbf{accessible}, as they can be retrieved without incurring any charges. 
Ensuring \textbf{interoperability}, the dataset is provided in a simple, standardized format (CSV for reviews and JPG for images). This facilitates their seamless integration.
Finally, the dataset is designed to be highly \textbf{re-usable}. The inclusion of a metadata.txt file serves to document information about the variables, offering comprehensive insights into the dataset's structure and context. 
In addition to the Zenodo repository at: \url{https://zenodo.org/records/10511456}, we release our work on GitHub at: \url{https://github.com/iamalegambetti/aigen-foodreview}, including trained unimodal and multimodal detectors. 

\section{Acknowledgments}
This work was funded by Fundação para a Ciência e a Tecnologia (UIDB/00124/2020, UIDP/00124/2020 and Social Sciences DataLab - PINFRA/22209/2016), POR Lisboa and POR Norte (Social Sciences DataLab, PINFRA/22209/2016).

\bibliography{aaai24}

\end{document}